\documentclass{article} % For LaTeX2e
\usepackage[accepted]{arxiv}
\usepackage{times}
\usepackage{amsmath}
\usepackage{amsthm}
\usepackage{amssymb}
\usepackage{graphicx}
\usepackage{xspace}
\usepackage{xcolor}
\usepackage{tabularx}
\usepackage{multicol}
\usepackage{multirow}
\usepackage{url}
\usepackage{natbib}
\usepackage{wrapfig}
\usepackage{comment}
\usepackage{listings}
\usepackage{color}
\usepackage[utf8]{inputenc}
\usepackage{fancyvrb}
\ifdefined\nohyperref\else\ifdefined\hypersetup
  \definecolor{mydarkblue}{rgb}{0,0.08,0.45}
  \hypersetup{ %
    pdftitle={},
    pdfauthor={},
    pdfsubject={Proceedings of the International Conference on Machine Learning 2015},
    pdfkeywords={},
    pdfborder=0 0 0,
    pdfpagemode=UseNone,
    colorlinks=true,
    linkcolor=mydarkblue,
    citecolor=mydarkblue,
    filecolor=mydarkblue,
    urlcolor=mydarkblue,
    pdfview=FitH}

  \ifdefined\isaccepted \else
    \hypersetup{pdfauthor={Anonymous Submission}}
  \fi
\fi\fi

\usepackage[normalem]{ulem}

\newcommand{\bmx}[0]{\begin{bmatrix}}
\newcommand{\emx}[0]{\end{bmatrix}}

\newcommand{\vect}[1]{\mathbf{#1}}
\newcommand{\vects}[1]{\boldsymbol{#1}}

\newcommand{\vc}[0]{\vect{c}}
\newcommand{\vh}[0]{\vect{h}}

\newcommand{\vx}[0]{\vect{x}}
\newcommand{\vz}[0]{\vect{z}}
\newcommand{\vw}[0]{\vect{w}}

\newcommand{\vf}[0]{\vect{f}}
\newcommand{\vi}[0]{\vect{i}}
\newcommand{\vo}[0]{\vect{o}}

\newcommand{\vu}[0]{\vect{u}}

\newcommand{\vr}[0]{\vect{r}}

\newcommand{\TT}[0]{\vects{\theta}}

\newcommand{\ts}{\rule{0pt}{2.6ex}}       % Top strut
\newcommand{\ms}{\rule{0pt}{0ex}}         % Middle strut
\newcommand{\bs}{\rule[-1.2ex]{0pt}{0pt}} % Bottom strut

\newcommand{\ImportantChars}[1]{\textbf{#1}}

\graphicspath{{./figures/}}

\begin{document}
%\maketitle
\twocolumn[
\icmltitle{Gated Feedback Recurrent Neural Networks}
\icmlauthor{Junyoung Chung}{junyoung.chung@umontreal.ca}
\icmlauthor{Caglar Gulcehre}{caglar.gulcehre@umontreal.ca}
\icmlauthor{Kyunghyun Cho}{kyunghyun.cho@umontreal.ca}
\icmlauthor{Yoshua Bengio$^\ast$}{find-me@the.web}
\icmladdress{Dept. IRO, Universit\'e de Montr\'eal, $^\ast$CIFAR Senior Fellow}
%\icmladdress{D\'epartement d'informatique et de recherche op\'erationnelle, Universit\'e de Montr\'eal}
%\icmladdress{Dept. IRO, Universit\'e de Montr\'eal. Montr\'eal (QC), H2C 3J7, Canada}

\icmlkeywords{deep learning, neural networks, recurrent neural networks, machine learning, ICML}
\vskip 0.3in
]
\begin{abstract}
    In this work, we propose a novel recurrent neural network (RNN) architecture.
    The proposed RNN, gated-feedback RNN (GF-RNN), extends
    the existing approach of stacking multiple recurrent layers by allowing
    and controlling signals flowing from upper recurrent layers to lower
    layers using a global gating unit for each pair of layers.
    The recurrent signals exchanged between layers are gated
    adaptively based on the previous hidden states and the current input. We
    evaluated the proposed GF-RNN with different types of recurrent units,
    such as $\tanh$, long short-term memory and gated recurrent units, on
    the tasks of character-level language modeling and Python program
    evaluation. Our empirical evaluation of different RNN units, revealed
    that in both tasks, the GF-RNN outperforms the conventional approaches
    to build deep stacked RNNs. We suggest that the improvement
    arises because the GF-RNN can adaptively assign different layers
    to different timescales and layer-to-layer interactions (including
    the top-down ones which are not usually present in a stacked RNN)
    by learning to gate these interactions.
\end{abstract}

\section{Introduction}
\label{sec:intro}

Recurrent neural networks (RNNs) have been widely studied and used for various
machine learning tasks which involve sequence modeling, especially when the input and
output have variable lengths. Recent studies have revealed that RNNs using gating units
can achieve promising results in both classification and generation tasks~\citep[see,
e.g.,][]{graves2013generating,Bahdanau-et-al-arxiv2014,sutskever2014sequence}.

Although RNNs can theoretically capture any long-term dependency in an input
sequence, it is well-known to be difficult to train an RNN to actually do
so~\citep{Hochreiter91,Bengio1994ITNN, hochreiter1998vanishing}. One of the most
successful and promising approaches to solve this issue is by modifying the RNN
architecture e.g., by using a gated activation function, instead of the usual
state-to-state transition function composing an affine transformation and a
point-wise nonlinearity.  A gated activation function, such as the long
short-term memory~\citep[LSTM,][]{Hochreiter+Schmidhuber-1997} and the gated
recurrent unit~\citep[GRU,][]{cho2014learning}, is designed to have more
persistent memory so that it can capture long-term dependencies more easily.

Sequences modeled by an RNN can contain both fast changing and slow changing
components, and these underlying components are often structured in a
hierarchical manner, which, as first pointed out by~\citet{el1995hierarchical} can
help to extend the ability of the RNN to learn to model longer-term dependencies.
A conventional way to encode this hierarchy in an RNN has
been to stack multiple levels of recurrent
layers~\citep{Schmidhuber92,el1995hierarchical,graves2013generating,hermans2013training}.
More recently, \citet{koutnik2014clockwork} proposed a more explicit approach to
partition the hidden units in an RNN into groups such that each group receives
the signal from the input and the other groups at a separate, predefined rate,
which allows feedback information between these partitions to be propagated at
multiple timescales. \citet{stollenga2014deep} recently showed the 
importance of feedback information across multiple levels of feature hierarchy,
however, with feedforward neural networks.

In this paper, we propose a novel design for RNNs, called a gated-feedback RNN
(GF-RNN), to deal with the issue of learning multiple adaptive timescales. The proposed
RNN has multiple levels of recurrent layers like stacked RNNs do. However, it uses
gated-feedback connections from upper recurrent layers to the lower ones. This makes
the hidden states across a pair of consecutive timesteps fully connected. To
encourage each recurrent layer to work at different timescales, the proposed
GF-RNN controls the strength of the temporal (recurrent) connection adaptively.
This effectively lets the model to adapt its structure based on the input sequence.

We empirically evaluated the proposed model against the conventional stacked RNN
and the usual, single-layer RNN on the task of language modeling and Python
program evaluation~\citep{zaremba2014learning}.
Our experiments reveal that the proposed model significantly outperforms
the conventional approaches on two different datasets.

\section{Recurrent Neural Network}
\label{sec:rnn}

An RNN is able to process a sequence of arbitrary length by recursively applying
 a transition function to its internal hidden states for each symbol of the input sequence.
The activation of the hidden states at timestep $t$ is computed as a function $f$ of the current input symbol
$\vx_t$ and the previous hidden states $\vh_{t-1}$:
\begin{align*}
    \vh_t =& f\left(\vx_t, \vh_{t-1}\right).
\end{align*}

It is common to use the state-to-state transition function $f$ as the composition of
an element-wise nonlinearity with an affine transformation of both $\vx_t$ and $\vh_{t-1}$:
\begin{align}
    \label{eq:rnn_hidden_trad}
    \vh_t =& \phi\left(W \vx_t + U \vh_{t-1}\right),
\end{align}
where $W$ is the input-to-hidden weight matrix, $U$ is the state-to-state recurrent weight matrix,
and $\phi$ is usually a logistic sigmoid function or a hyperbolic tangent function.

We can factorize the probability of a sequence of arbitrary length into
\begin{align*}
    p(x_1, \cdots, x_{\scriptscriptstyle{T}}) = p(x_1) p(x_2 \mid x_1) \cdots p(x_{\scriptscriptstyle{T}} \mid x_1, \cdots,
    x_{\scriptscriptstyle{T-1}}).
\end{align*}
Then, we can train an RNN to model this distribution by letting it predict the probability
of the next symbol $x_{t+1}$ given hidden states $\vh_t$ which is a function of all the
previous symbols $x_1, \cdots, x_{t-1}$ and current symbol $x_t$:
\begin{align*}
    p(x_{t+1} \mid x_1, \cdots, x_t) = g\left(\vh_{t} \right).
\end{align*}
This approach of using a neural network to model a probability distribution over sequences
is widely used, for instance, in language modeling~\citep[see,
e.g.,][]{BenDucVin01-short,Mikolov-thesis-2012}.

\subsection{Gated Recurrent Neural Network}

The difficulty of training an RNN to capture long-term dependencies has been
known for long~\citep{Hochreiter91,Bengio1994ITNN,hochreiter1998vanishing}. A previously successful
approaches to this fundamental challenge has been to modify the state-to-state transition function
to encourage some hidden units to adaptively maintain long-term memory, creating paths in the
time-unfolded RNN, such that gradients can flow over many timesteps.

Long short-term memory (LSTM) was proposed by
\citet{Hochreiter+Schmidhuber-1997} to specifically address this issue of
learning long-term dependencies. The LSTM maintains a separate memory cell
inside it that updates and exposes its content only when deemed necessary. More
recently, \citet{cho2014learning} proposed a gated recurrent unit (GRU) which
adaptively remembers and forgets its state based on the input signal to the
unit. Both of these units are central to our proposed model, and we will
describe them in more details in the remainder of this section.

\subsubsection{Long Short-Term Memory}
\label{sec:lstm}

Since the initial 1997 proposal, several variants of the LSTM have been
introduced~\cite{Gers2000,zaremba2014}. Here we follow the implementation provided by
\citet{zaremba2014}.

Such an LSTM unit consists of a memory cell $c_t$, an {\it input} gate $i_t$,
a {\it forget} gate $f_t$, and an {\it output} gate $o_t$. The memory cell carries
the memory content of an LSTM unit, while the gates control the amount of
changes to and exposure of the memory content.
The content of the memory cell $c_t^j$ of the $j$-th LSTM unit at timestep $t$
is updated similar to the form of a gated leaky neuron, i.e.,
as the weighted sum of the new content $\tilde{c}_t^j$ and the previous
memory content $c_{t-1}^j$ modulated by the input and forget gates,
$i_t^j$ and $f_t^j$, respectively:
\begin{align}
    \label{eq:lstm_cell}
    c_t^j = f_t^j c_{t-1}^j + i_t^j \tilde{c}_t^j,
\end{align}
where
\begin{align}
    \label{eq:lstm_new_cell}
    \tilde{\vc}_t = \tanh\left(W_c \vx_t + U_c \vh_{t-1}\right).
\end{align}
The input and forget gates control how much new content should be
{\it memorized} and how much old content should be {\it forgotten},
respectively. These gates are computed from the previous hidden
states and the current input:
\begin{align}
    \label{eq:lstm_input}
    \vi_t =& \sigma\left(W_i\vx_t + U_i\vh_{t-1}\right),\\
    \label{eq:lstm_forget}
    \vf_t =& \sigma\left(W_f\vx_t + U_f\vh_{t-1}\right),
\end{align}
where $\vi_t=\left[ i_t^k \right]_{k=1}^p$ and $\vf_t=\left[ f_t^k
\right]_{k=1}^p$ are respectively the vectors of the input and forget
gates in a recurrent layer composed of $p$ LSTM units. $\sigma(\cdot)$
is an element-wise logistic sigmoid function. $\vx_t$ and $\vh_{t-1}$ are the
input vector and previous hidden states of the LSTM units, respectively.

Once the memory content of the LSTM unit is updated, the hidden state
$h_t^j$ of the $j$-th LSTM unit is computed as:
\begin{align*}
    h_t^j = o_t^j \tanh\left( c_t^j \right).
\end{align*}
The output gate $o_t^j$ controls to which degree the memory content is
exposed. Similarly to the other gates, the output gate also depends on
the current input and the previous hidden states such that
\begin{align}
    \label{eq:lstm_output}
    \vo_t = \sigma\left(W_o\vx_t + U_o\vh_{t-1}\right).
\end{align}

In other words, these gates and the memory cell allow an LSTM unit to adaptively
{\it forget}, {\it memorize} and {\it expose} the memory content. If the
detected feature, i.e., the memory content, is deemed important, the
forget gate will be closed and carry the memory content across many timesteps,
which is equivalent to capturing a long-term dependency. On the other hand, the
unit may decide to reset the memory content by opening the forget gate.
Since these two modes of operations can happen simultaneously across different
LSTM units, an RNN with multiple LSTM units may capture both fast-moving and slow-moving components.

\subsubsection{Gated Recurrent Unit}
\label{sec:gru}

The GRU was recently proposed by \citet{cho2014learning}. Like the LSTM,
it was designed to adaptively {\it reset} or {\it update} its memory
content. Each GRU thus has a {\it reset} gate $r_t^j$ and an {\it update} gate
$z_t^j$ which are reminiscent of the forget and input
gates of the LSTM. However, unlike the LSTM, the GRU fully exposes its memory
content each timestep and balances between the previous memory content and the
new memory content strictly using leaky integration, albeit with its
adaptive time constant controlled by update gate $z_t^j$.

At timestep $t$, the state $h_t^j$ of the $j$-th GRU is computed by
\begin{align}
    \label{eq:gru_memory_up}
    h_t^j = (1 - z_t^j) h_{t-1}^j + z_t^j \tilde{h}_t^j,
\end{align}
where $h_{t-1}^j$ and $\tilde{h}_t^j$ respectively correspond to the previous memory content
and the new candidate memory content. The update
gate $z_t^j$ controls how much of the previous memory content is to be forgotten and
how much of the new memory content is to be added. The update gate is computed based on the
previous hidden states $\vh_{t-1}$ and the current input $\vx_t$:
\begin{align}
    \label{eq:gru_update_gate}
   \vz_t =& \sigma\left(W_z \vx_t + U_z\vh_{t-1}\right),
\end{align}
The new memory content $\tilde{h}_t^j$ is computed similarly to the conventional
transition function in Eq.~\eqref{eq:rnn_hidden_trad}:
\begin{align}
    \label{eq:gru_new_act}
    \tilde{\vh}_t = \tanh\left(W \vx_t + \vr_t \odot U\vh_{t-1}
    \right),
\end{align}
where $\odot$ is an element-wise multiplication.

One major difference from the traditional transition function
(Eq.~\eqref{eq:rnn_hidden_trad}) is that the states of the previous step
$\vh_{t-1}$ is modulated by the reset gates $\vr_t$. This behavior allows a
GRU to ignore the previous hidden states whenever it is deemed necessary
considering the previous hidden states and the current input:
\begin{align}
    \label{eq:gru_memory_reset}
   \vr_t =& \sigma\left(W_r \vx_t + U_r\vh_{t-1}\right).
\end{align}

The update mechanism helps the GRU to capture long-term dependencies. Whenever a
previously detected feature, or the memory content is considered to be important
for later use, the update gate will be closed to carry the current
memory content across multiple timesteps. The reset mechanism helps the GRU
to use the model capacity efficiently by allowing it to reset whenever
 the detected feature is not necessary anymore.

\section{Gated Feedback Recurrent Neural Network}
\label{sec:gfrnn}

Although capturing long-term dependencies in a sequence is an important and
difficult goal of RNNs, it is worthwhile to notice
that a sequence often consists of both slow-moving and fast-moving components,
of which only the former corresponds to long-term dependencies. Ideally, an RNN
needs to capture both long-term and short-term dependencies.

\citet{el1995hierarchical} first showed that an RNN
can capture these dependencies of different timescales more easily and
efficiently when the hidden units of the RNN is explicitly partitioned into
groups that correspond to different timescales. The clockwork RNN (CW-RNN)
\citep{koutnik2014clockwork} implemented this
by allowing the $i$-th module to operate at the rate of $2^{i-1}$, where $i$ is
a positive integer, meaning that the module is updated only when $t \mbox{~mod~} 2^{i-1}=0$.
This makes each module to operate at different rates.
In addition, they precisely defined the
connectivity pattern between modules by allowing the $i$-th module to be
affected by $j$-th module when $j>i$.

\begin{figure*}[ht]
    \centering
    \begin{minipage}[b]{0.48\textwidth}
        \centering
        \includegraphics[width=0.98\columnwidth]{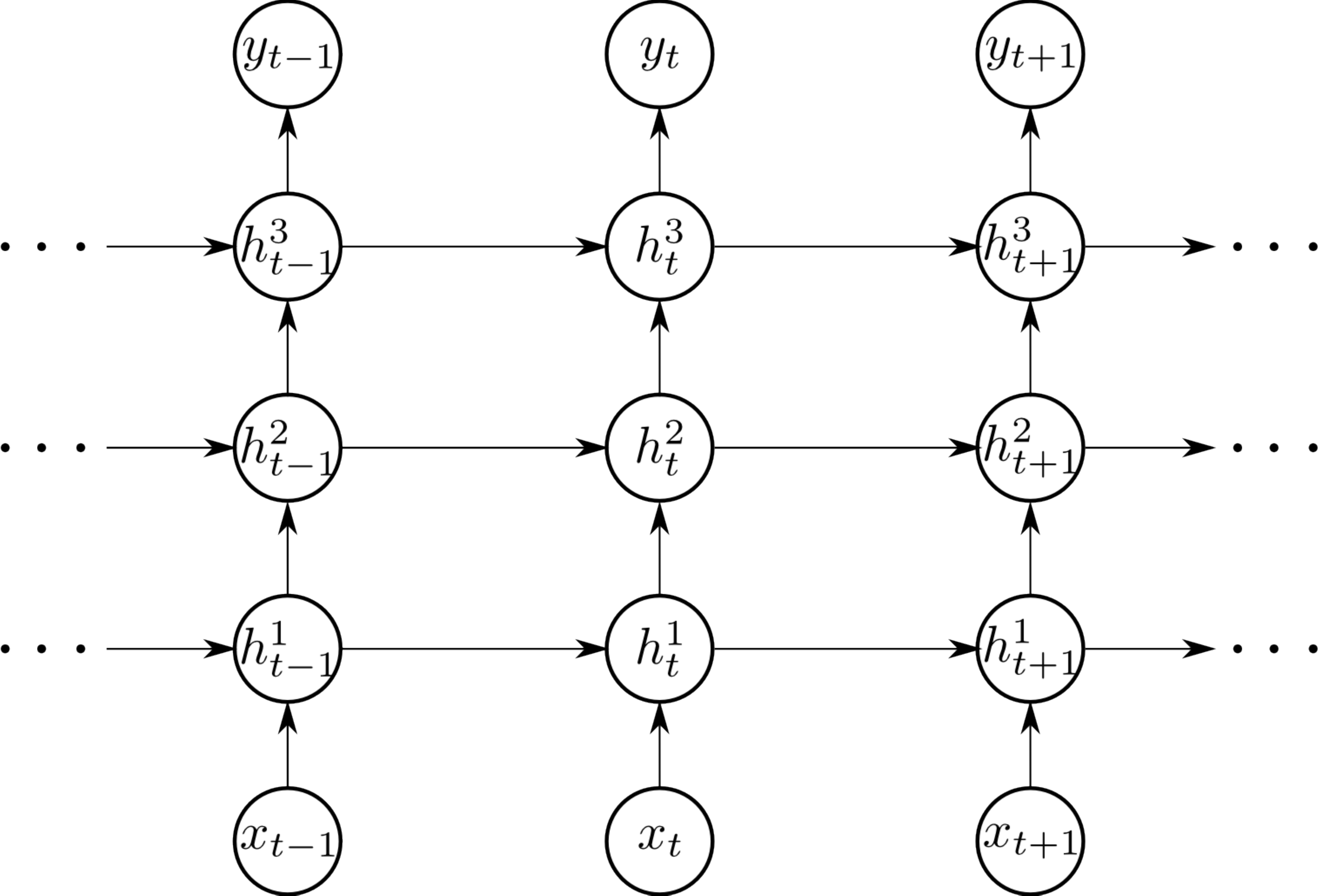}
    \end{minipage}
    \hfill
    \begin{minipage}[b]{0.48\textwidth}
        \centering
        \includegraphics[width=0.98\columnwidth]{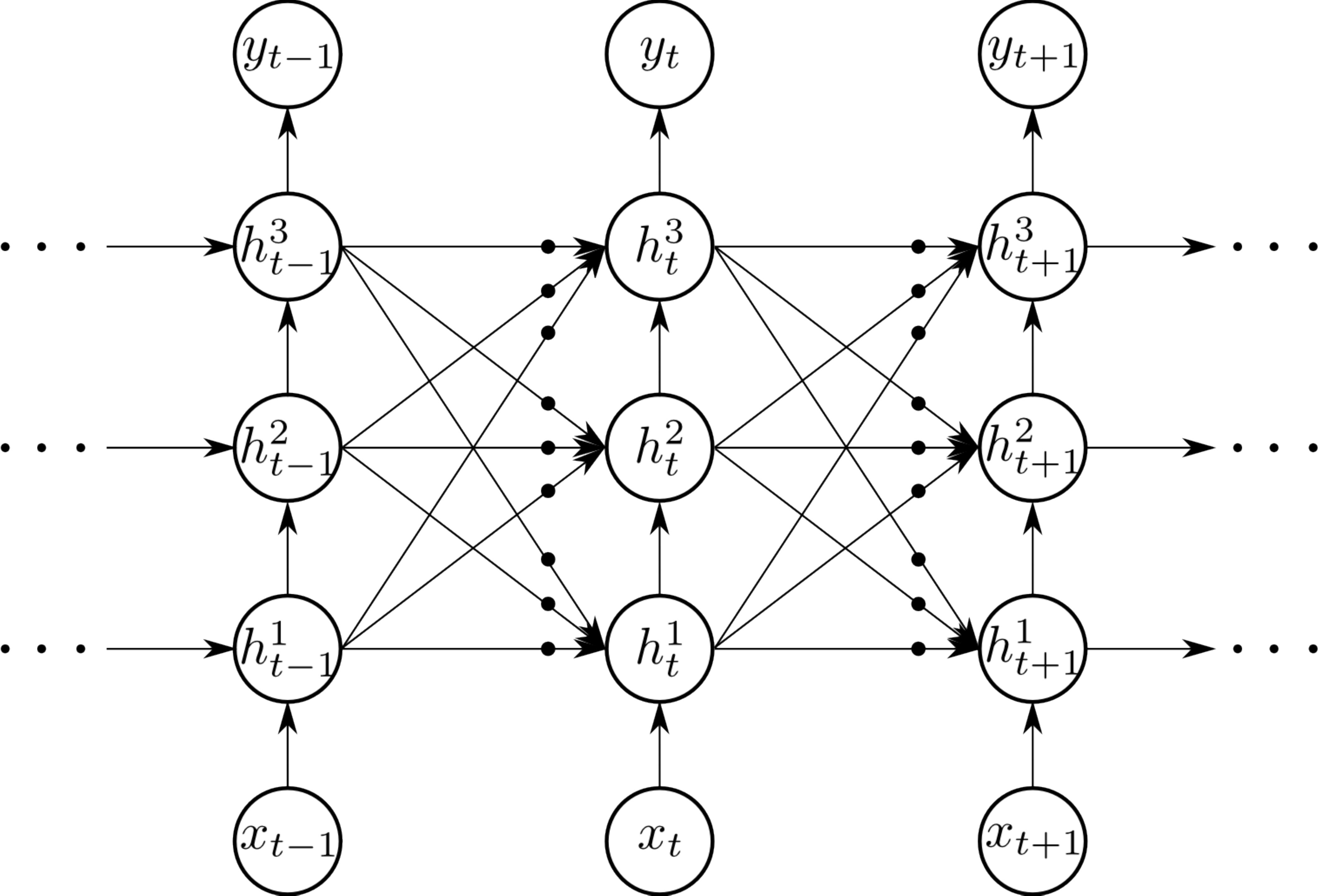}
    \end{minipage}

    \vspace{2mm}
    \begin{minipage}{0.48\textwidth}
        \centering
        (a) Conventional stacked RNN
    \end{minipage}
    \hfill
    \begin{minipage}{0.48\textwidth}
        \centering
        (b) Gated Feedback RNN
    \end{minipage}

    \caption{
        Illustrations of (a) conventional stacking approach and
        (b) gated-feedback approach to form a deep RNN architecture.
        Bullets in (b) correspond to global reset gates.
        Skip connections are omitted to simplify the visualization of networks.
    }
    \label{fig:stacking_scheme}
\end{figure*}

Here, we propose to generalize the CW-RNN by allowing the model to adaptively
adjust the connectivity pattern between the hidden layers in the consecutive
timesteps.  Similar to the CW-RNN, we partition the hidden units into multiple
modules in which each module corresponds to a different layer in a stack of
recurrent layers.

Unlike the CW-RNN, however, we do not set an explicit rate for each module.
Instead, we let each module operate at different timescales by
hierarchically stacking them. Each module is fully connected to all the
other modules across the stack and itself. In other words, we do not define the
connectivity pattern across a pair of consecutive timesteps. This is contrary
to the design of CW-RNN and the conventional stacked RNN.  The recurrent
connection between two modules, instead, is gated by a logistic unit ($\left[ 0,
1\right]$) which is computed based on the current input and the previous
states of the hidden layers.  We call this gating unit a {\it global reset}
gate, as opposed to a unit-wise reset gate which applies only to a single unit (See
Eqs.~\eqref{eq:lstm_cell}~and~\eqref{eq:gru_new_act}).

The global reset gate is computed as:
\[
    %g^{i \to j} = \sigma\left( \vw^{j-1 \to j}_g \; \vh_t^{j-1} + \sum_{i'=1}^L\vu^{i' \to j}_g \;
    %\vh^{i'}_{t-1} \right),
    g^{i \to j} = \sigma\left( \vw^{i \to j}_g \; \vh_t^{j-1} + \vu_g^{i \to j}\
                   \vh_{t-1}^{\ast} \right),
               \]
%change_2 Reviewer_3 asked to make this part more clear
where $\vh_{t-1}^{\ast}$ is the concatenation of all the hidden states from the
previous timestep $t-1$.  The superscript $^{i \to j}$ is an index of
associated set of parameters for the transition from layer $i$ in timestep $t-1$
to layer $j$ in timestep $t$.  $\vw_g^{i \to j}$ and $\vu_g^{i \to j}$ are
respectively the weight vectors for the current input and the previous hidden
states. When $j=1$, $\vh_t^{j-1}$ is $\vx_t$.

In other words, the signal from $\vh^i_{t-1}$ to $\vh^j_t$ is controlled by a
single scalar $g^{i \to j}$ which depends on the input $\vx_t$ and all the
previous hidden states $\vh_{t-1}^{\ast}$.

We call this RNN with a fully-connected recurrent transitions and global reset
gates, a {\it gated-feedback RNN} (GF-RNN). Fig.~\ref{fig:stacking_scheme}
illustrates the difference between the conventional stacked RNN and our proposed
GF-RNN. In both models, information flows from lower recurrent layers to upper
recurrent layers.  The GF-RNN, however, further allows information
from the upper recurrent layer, corresponding to coarser timescale, flows back
into the lower recurrent layers, corresponding to finer timescales.

In the remainder of this section,
we describe how to use the previously described LSTM unit, GRU, and more
traditional $\tanh$ unit in the GF-RNN.

\subsection{Practical Implementation of GF-RNN}

\paragraph{$\tanh$ Unit.}

For a stacked $\tanh$-RNN, the signal from the previous timestep is gated.
The hidden state of the $j$-th layer is computed by
\begin{align*}
    \vh_t^j =& \tanh\left(W^{j-1\to j} \vh_t^{j-1} +
    \sum_{i=1}^L g^{i \to j} U^{i \to j}  \vh_{t-1}^i\right),
\end{align*}
where $L$ is the number of hidden layers, $W^{j-1\to j}$ and $U^{i \to j}$ are
the weight matrices of the current input and the previous hidden states of the $i$-th module,
respectively. Compared to Eq.~\eqref{eq:rnn_hidden_trad}, the only difference is that
 the previous hidden states are from multiple layers and controlled by the global reset gates.

\paragraph{Long Short-Term Memory and Gated Recurrent Unit.}

In the cases of LSTM and GRU, we do not use the global reset gates when
computing the unit-wise gates. In other words,
Eqs.~\eqref{eq:lstm_input}--\eqref{eq:lstm_output} for LSTM, and
Eqs.~\eqref{eq:gru_update_gate} and \eqref{eq:gru_memory_reset} for GRU
are not modified. We only use the global reset gates when computing the new state
(see Eq.~\eqref{eq:lstm_new_cell} for LSTM, and Eq.~\eqref{eq:gru_new_act} for
GRU).

The new memory content of an LSTM at the $j$-th
layer is computed by
\begin{align*}
    \tilde{\vc}_t^j = \tanh\left(W_c^{j-1\to j} \vh_t^{j-1} + \sum_{i=1}^L g^{i \to j} U_c^{i \to j}\vh_{t-1}^i\right).
\end{align*}
In the case of a GRU, similarly,
\begin{align*}
    \tilde{\vh}_t^j = \tanh\left(W^{j-1 \to j} \vh_t^{j-1} + \vr_t^j \odot \sum_{i=1}^L g^{i \to j} U^{i \to j} \vh_{t-1}^i
    \right).
\end{align*}

\section{Experiment Settings}

\subsection{Tasks}

We evaluated the proposed GF-RNN on character-level language
modeling and Python program evaluation. Both tasks are representative examples
of discrete sequence modeling, where a model is trained to minimize the negative
log-likelihood of training sequences:
%\vspace{-0.15in}
\begin{align*}
  \min_{\TT} \frac{1}{N} \sum_{n=1}^N \sum_{t=1}^{T_n} -\log p\left(x_t^n
        \mid x_1^n, \dots, x_{t-1}^n; \TT \right),
\end{align*}
where $\TT$ is a set of model parameters.

\subsubsection{Language Modeling}

We used the dataset made available as a part of the human knowledge compression
contest~\citep{Hutter2012}. We refer to this dataset as the {\it Hutter
dataset}. The dataset, which was built from English Wikipedia, contains $100$
MBytes of characters which include Latin alphabets, non-Latin alphabets, XML
markups and special characters. Closely following the protocols in
\citep{mikolov2012subword,graves2013generating}, we used the first $90$ MBytes of
characters to train a model, the next $5$ MBytes as a validation set, and the
remaining as a test set, with the vocabulary of 205 characters including a token
for an unknown character. We used the average number of bits-per-character (BPC, $E[-\log_2
P(x_{t+1}|\vh_t)]$) to measure the performance of each model on the Hutter dataset.

\subsubsection{Python Program Evaluation}

\citet{zaremba2014learning} recently showed that an RNN, more
specifically a stacked LSTM, is able to execute a short Python script. Here, we
compared the proposed architecture against the conventional stacking approach
model on this task, to which refer as {\it Python program evaluation}.

Scripts used in this task include addition, multiplication, subtraction,
for-loop, variable assignment, logical comparison and if-else statement.
The goal is to generate, or predict, a correct return value of a given Python
script.
The input is a program while the output is the result of a print statement:
every input script ends with a print statement.
Both the input script and the output are sequences of characters, where
the input and output vocabularies respectively consist of $41$ and $13$ symbols.

The advantage of evaluating the models with this task is that we can
artificially control the difficulty of each sample (input-output pair). The
difficulty is determined by the number of nesting levels in the input sequence and
the length of the target sequence. We can do a finer-grained analysis of
each model by observing its behavior on examples of different difficulty levels.

%change_5 Details on python program evaluation test procedure
In Python program evaluation, we closely follow \citep{zaremba2014learning} and
compute the test accuracy as the next step symbol prediction given a sequence
of correct preceding symbols.
%change_5 end

\subsection{Models}
\label{sec:models}

We compared three different RNN architectures:
a single-layer RNN, a stacked RNN and the proposed GF-RNN.
For each architecture, we evaluated three different transition functions: $\tanh$ + affine, long
short-term memory (LSTM) and gated recurrent unit (GRU). For fair comparison,
we constrained the number of parameters of each model to be roughly similar to each other.

For each task, in addition to these capacity-controlled experiments, we conducted
a few extra experiments to further test and better understand the properties of
the GF-RNN.

\begin{table}[t]
    \caption{The sizes of the models used in character-level language modeling.
        Gated Feedback L is a GF-RNN with a same number of hidden units as a Stacked RNN
        (but more parameters). The number of units is shown
        as {\tt (number of hidden layers)} $\times$ {\tt (number of hidden units per layer)}.}
    \label{tab:hutter_cap}

    \centering
    \begin{tabular}{c | c | c}
        Unit & Architecture & \# of Units \\
        \hline
        \hline
        \multirow{3}{*}{$\tanh$}
        & Single & $1 \times 1000$ \ts\\
        & Stacked & $3 \times 390$ \ms\\
        & Gated Feedback & $3 \times 303$ \bs\\
        \hline
        \multirow{4}{*}{GRU}
        & Single & $1 \times 540$ \ts\\
        & Stacked & $3 \times 228$ \ms\\
        & Gated Feedback & $3 \times 165$ \ms\\
        & Gated Feedback L & $3 \times 228$ \bs\\
        \hline
        \multirow{4}{*}{LSTM}
        & Single & $1 \times 456$ \ts\\
        & Stacked & $3 \times 191$ \ms\\
        & Gated Feedback & $3 \times 140$ \ms\\
        & Gated Feedback L& $3 \times 191$ \bs\\
        \hline
    \end{tabular}

    \vskip -10mm
\end{table}

\begin{figure*}[ht]
    \begin{minipage}{1\textwidth}
        \centering
        \begin{minipage}[b]{0.45\textwidth}
            \centering
            \includegraphics[width=1.\textwidth,clip=true]{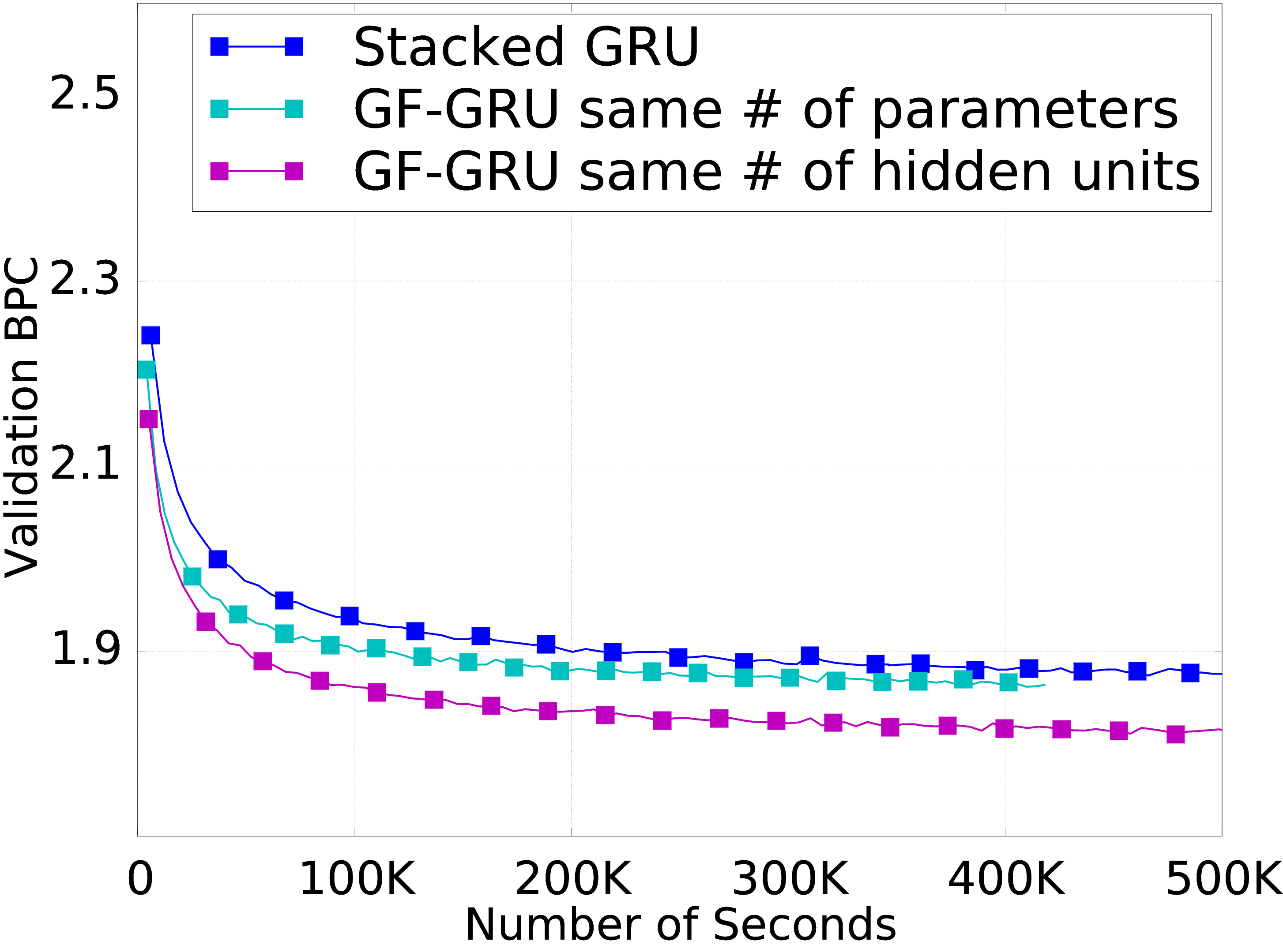}
        \end{minipage}
        \hfill
        \begin{minipage}[b]{0.45\textwidth}
            \centering
            \includegraphics[width=1.\textwidth,clip=true]{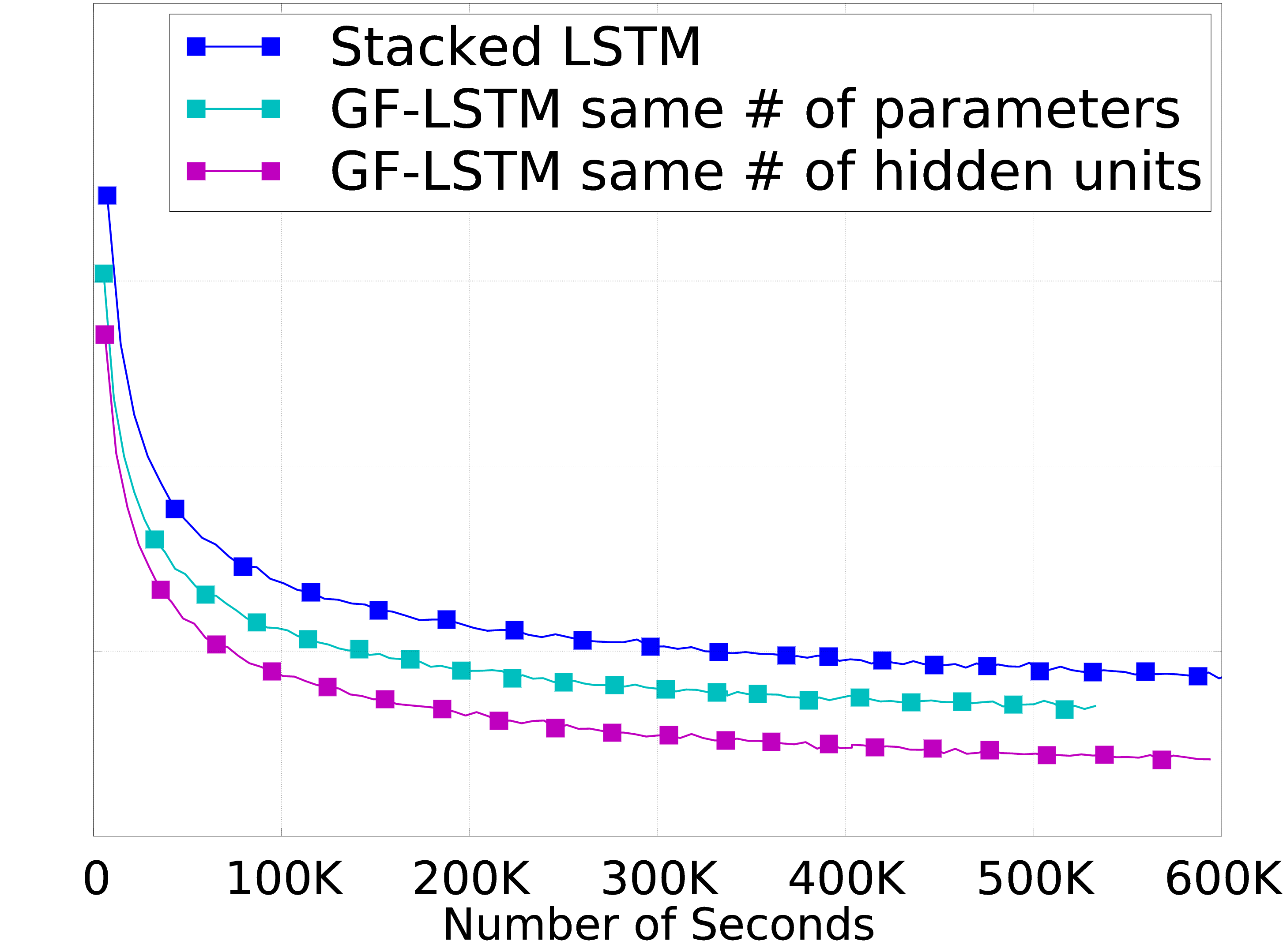}
        \end{minipage}

        \begin{minipage}{0.45\textwidth}
            \centering
            (a) GRU
        \end{minipage}
        \hfill
        \begin{minipage}{0.45\textwidth}
            \centering
            (b) LSTM
        \end{minipage}

    \end{minipage}
    \caption{Validation learning curves of three different RNN architectures;
    Stacked RNN, GF-RNN with the same number of model parameters and
GF-RNN with the same number of hidden units. The curves represent training up
to $100$ epochs. Best viewed in colors.}
    \label{fig:hutter_results}
\end{figure*}

\subsubsection{Language Modeling}

For the task of character-level language modeling, we constrained the number of
parameters of each model to correspond to that of a single-layer RNN with $1000$
$\tanh$ units (see Table~\ref{tab:hutter_cap} for more details). Each model is
trained for at most $100$ epochs.

We used RMSProp~\citep{Hinton-Coursera2012} and momentum to tune the model
parameters \citep{graves2013generating}. According to the preliminary experiments and their results on the
validation set, we used a learning rate of $0.001$ and momentum coefficient of
$0.9$  when training the models having either GRU or LSTM units. It was
necessary to choose a much smaller learning rate of $5 \times 10^{-5}$ in the
case of $\tanh$ units to ensure the stability of learning. Whenever the norm of
the gradient explodes, we halve the learning rate.

Each update is done using a minibatch of $100$ subsequences of length $100$ each,
to avoid memory overflow problems when unfolding in time for backprop. We
approximate full back-propagation by carrying the hidden states computed at the
previous update to initialize the hidden units in the next update. After every
$100$-th update, the hidden states were reset to all zeros.

\begin{table}[ht]
    \caption{Test set BPC (lower is better) of models trained on the Hutter dataset for a $100$ epochs.
             ($\ast$) The gated-feedback RNN with the global reset gates
     fixed to $1$ (see Sec.~\ref{sec:language_modeling} for details).
     Bold indicates statistically significant winner over the column (same type of units, different
     overall architecture).}
    \label{tab:hutter_results}
    \vskip 0.15in
    \centering
    \begin{tabular}{ c | c | c | c }
        ~ & $\tanh$ & GRU & LSTM \\
        \hline
        \hline
        Single-layer & 1.937 & 1.883 & 1.887 \\
        \cline{1-4}
        Stacked & \bf 1.892 & 1.871 & 1.868 \\
        \cline{1-4}
        Gated Feedback & 1.949 & \bf 1.855 & \bf 1.842 \\
        \cline{1-4}
        Gated Feedback L & -- & \bf 1.813 & \bf 1.789 \\
        \cline{1-4}
        Feedback$^{\ast}$ & -- & -- & 1.854 \\
        \cline{1-4}
        \hline
    \end{tabular}
    \vspace{-0.1in}
\end{table}

\begin{comment}
\begin{figure}[h]
    \centering
        \includegraphics[width=0.45\textwidth,clip=true]{valid_curves_by_group_lstm_gate.png}
    \caption{Validation learning curves of the stacked LSTM, GF-LSTM with the
    global reset gates and GF-LSTM without them. Best viewed in colors.}
    \label{fig:global_reset_gates}
\end{figure}
\end{comment}

\subsubsection{Python Program Evaluation}

For the task of Python program evaluation, we used an RNN encoder-decoder based approach
to learn the mapping from Python scripts to the corresponding outputs as done by
\citet{cho2014learning,sutskever2014sequence} for machine translation.
When training the models, Python scripts are fed into the encoder RNN, and the hidden state
of the encoder RNN is unfolded for $50$ timesteps. Prediction is performed by the decoder RNN whose
initial hidden state is initialized with the last hidden state of the encoder RNN.
The first hidden state of encoder RNN $h_0$ is always initialized to a zero vector.

For this task, we used GRU and LSTM units either with or without the gated-feedback connections.
%We constrained the number of parameters to $2.4$MBytes to control the capacity of each model.
Each encoder or decoder RNN has three hidden layers.
For GRU, each hidden layer contains $230$ units, and for LSTM each hidden layer contains
$200$ units.

Following \citet{zaremba2014learning}, we used the {\it mixed}
curriculum strategy for training each model, where each training example has a
random difficulty sampled uniformly. We generated $320,000$ examples using the
script provided by \citet{zaremba2014learning}, with the nesting randomly sampled from
$\left[ 1, 5\right]$ and the target length from $\left[1, 10^{10} \right]$.

We used Adam~\citep{kingma2014adam} to train our models, and each update was
using a minibatch with $128$ sequences. We used a learning rate of $0.001$ and
$\beta_1$ and $\beta_2$ were both set to $0.99$.
We trained each model for $30$ epochs, with early stopping
based on the validation set performance to prevent over-fitting.

At test time, we evaluated each model on multiple sets of test examples
where each set is generated using a fixed target length and number of nesting levels.
Each test set contains $2,000$ examples which are ensured not to overlap with
the training set.

\section{Results and Analysis}
\label{sec:results}
\begin{comment}
\begin{figure*}[ht]
    \begin{minipage}{1\textwidth}
        \centering
        \begin{minipage}[b]{0.48\textwidth}
            \centering
            \includegraphics[width=1.\textwidth,clip=true]{valid_curves_by_group_gru_time.png}
        \end{minipage}
        \hfill
        \begin{minipage}[b]{0.48\textwidth}
            \centering
            \includegraphics[width=1.\textwidth,clip=true]{valid_curves_by_group_lstm_time.png}
        \end{minipage}

        \begin{minipage}{0.48\textwidth}
            \centering
            (a) GRU
        \end{minipage}
        \hfill
        \begin{minipage}{0.48\textwidth}
            \centering
            (b) LSTM
        \end{minipage}

    \end{minipage}
    \caption{Validation learning curves of three different RNN architectures;
    Stacked RNN, GF-RNN with the same number of model parameters, and
GF-RNN with the same number of hidden units. The curves represent training up
to $100$ epochs. Best viewed in colors.}
    \label{fig:hutter_results}
\end{figure*}
\end{comment}

\begin{table*}[ht]
    \begin{minipage}{1\textwidth}
    \caption{Generated texts with our trained models.
             Given the seed at the left-most column (bold-faced font),
             the models predict next $200\sim300$ characters. Tabs, spaces and new-line characters
             are also generated by the models.}
    \label{tab:text_generation}
    \vskip 0.15in
    \centering
    \begin{tabular}{ | m{4cm} | m{5.5cm} | m{5.5cm} | }
        \hline
        Seed & Stacked LSTM & GF-LSTM \\
        \hline
\tiny
\begin{Verbatim}[commandchars=\\\{\}]
\ImportantChars{[[pl:Icon]]}
\ImportantChars{[[pt:Icon]]}
\ImportantChars{[[ru:Icon]]}
\ImportantChars{[[sv:Programspraket Icon]]</text>}
\ImportantChars{    </revision>}
\ImportantChars{  </page>}
\ImportantChars{  <page>}
\ImportantChars{    <title>Iconology</title>}
\ImportantChars{    <id>14802</id>}
\ImportantChars{    <revi}
\end{Verbatim}
\normalsize
&
\tiny
\begin{Verbatim}[commandchars=\\\{\}]
    \ImportantChars{<revi}sion>
    <id>15908383</id>
    <timestamp>
      2002-07-20T18:33:34Z
    </timestamp>
    <contributor>
      <username>The Courseichi</userrand
                vehicles in [[enguit]].
==The inhibitors and alphabetsy and moral/
hande in===In four [[communications]] and
\end{Verbatim}
\normalsize
&
\tiny
\begin{Verbatim}[commandchars=\\\{\}]
    \ImportantChars{<revi}sion>
    <id>41968413</id>
    <timestamp>
      2006-09-03T11:38:06Z
    </timestamp>
    <contributor>
      <username>Navisb</username>
      <id>46264</id>
    </contributor>
    <comment>The increase from the time
\end{Verbatim}
\normalsize
\\
        %\hline
        \cline{1-3}
\tiny
\begin{Verbatim}[commandchars=\\\{\}]
\ImportantChars{<title>Inherence relation</title>}
\ImportantChars{<id>14807</id>}
\ImportantChars{<revision>}
\ImportantChars{  <id>34980694</id>}
\ImportantChars{  <timestamp>}
\ImportantChars{    2006-01-13T04:19:25Z}
\ImportantChars{  </timestamp>}
\ImportantChars{  <contributor>}
\ImportantChars{    <username>Ro}
\end{Verbatim}
\normalsize
&
\tiny
\begin{Verbatim}[commandchars=\\\{\}]
       \ImportantChars{<username>Ro}bert]]
[[su:20 aves]]
[[vi:10 Februari]]
[[bi:16 agostoferosín]]
[[pt:Darenetische]]
[[eo:Hebrew selsowen]]
[[hr:2 febber]]
[[io:21 februari]]
[[it:18 de februari]]
\end{Verbatim}
\normalsize
&
\tiny
\begin{Verbatim}[commandchars=\\\{\}]
        \ImportantChars{<username>Ro}ma</username>
        <id>48</id>
      </contributor>
      <comment>Vly''' and when one hand
is angels and [[ghost]] borted and
''mask r:centrions]], [[Afghanistan]],
[[Glencoddic tetrahedron]], [[Adjudan]],
[[Dghacn]], for example, in which materials
dangerous (carriers) can only use with one
\end{Verbatim}
\normalsize
\\
        %\hline
        \cline{1-3}
    \end{tabular}
    %\vskip -0.1in
    \vskip -0.3in
    \end{minipage}
\end{table*}

\subsection{Language Modeling}
\label{sec:language_modeling}

It is clear from Table~\ref{tab:hutter_results} that the proposed gated-feedback
architecture outperforms the other baseline architectures that we have tried
when used together with widely used gated units such as LSTM and GRU. However,
the proposed architecture failed to improve the performance of a vanilla-RNN
with $\tanh$ units.
In addition to the final modeling performance, in Fig.~\ref{fig:hutter_results},
we plotted the learning curves of some models against wall-clock time (measured in
seconds). RNNs that are trained with the proposed gated-feedback architecture tends to
make much faster progress over time. This behavior is observed both when the number
of parameters is constrained and when the number of hidden units is constrained.
This suggests that the proposed GF-RNN significantly facilitates optimization/learning.

\textbf{Effect of Global Reset Gates}

After observing the superiority of the proposed gated-feedback architecture over the
single-layer or conventional stacked ones, we further trained another GF-RNN
with LSTM units, but this time, after fixing the global reset gates to $1$ to
validate the need for the global reset gates. Without the global reset gates,
feedback signals from the upper recurrent layers influence the lower recurrent
layer fully without any control.
%change_3 Removed the graph
The test set BPC of GF-LSTM without global reset gates was $1.854$ which is in
between the results of conventional stacked LSTM and GF-LSTM with global reset
gates (see the last row of Table~\ref{tab:hutter_results}) which confirms the
importance of adaptively gating the feedback connections.

%In Fig.~\ref{fig:global_reset_gates}, it can be
%seen that this omission of the global reset gates hurts the performance ({\color{cyan} cyan})
%compared to the one with the global reset gates ({\color{magenta} magenta}).
%The test set BPC of GF-LSTM without global reset gates was $1.854$.

%\begin{figure}[h]
%    \centering
%        \includegraphics[width=0.43\textwidth,clip=true]{./figures/valid_curves_by_group_lstm_gate.pdf}
%    \caption{Validation learning curves of the stacked LSTM, GF-LSTM with the global
%    reset gates and GF-LSTM without them. Best viewed in colors.}
%    \label{fig:global_reset_gates}
%\end{figure}
%change_3 end

\textbf{Qualitative Analysis: Text Generation}

Here we qualitatively evaluate the stacked LSTM and GF-LSTM trained earlier by
generating text. We choose a subsequence of characters from the test set and use
it as an initial seed.  Once the model finishes reading the seed text, we let
the model generate the following characters by sampling a symbol from {\it
softmax} probabilities of a timestep and then provide the symbol as next input.

%change_4 Add more details on what reviewer_2 asked
%We sampled $10$ times using different random seeds, but fixed initial seed
%snippets from the test set.  
Given two seed snippets selected randomly from the test set, we generated the
sequence of characters ten times for each model (stacked LSTM and GF-LSTM). We
show one of those ten generated samples per model and per seed snippet in
Table~\ref{tab:text_generation}.  We observe that the stacked LSTM failed to
close the tags with \small\texttt{</username>}\normalsize\hspace{0.5mm} and
\small\texttt{</contributor>}\normalsize \hspace{0.5mm} in both trials.
However, the GF-LSTM succeeded to close both of them, which shows that it
learned about the structure of XML tags. This type of behavior could be seen
throughout all ten random generations.
%change_4 end

\begin{table}[h]
    \caption{Test set BPC of neural language models trained on the Hutter
    dataset, MRNN = multiplicative RNN results from~\citet{sutskever2011generating}
    and Stacked LSTM results from~\citet{graves2013generating}.}
    \label{tab:hutter_results_large}
    \vskip 0.15in
    \centering
    \begin{tabular}{ c | c | c }
        MRNN & Stacked LSTM & GF-LSTM \\
        \hline
        \hline
        1.60 & 1.67 & {\bf 1.58} \\
        \cline{1-3}
        \hline
    \end{tabular}
\end{table}

\begin{figure*}[ht]
    \centering
    \begin{minipage}[b]{0.32\textwidth}
        \centering
        \includegraphics[width=0.98\columnwidth]{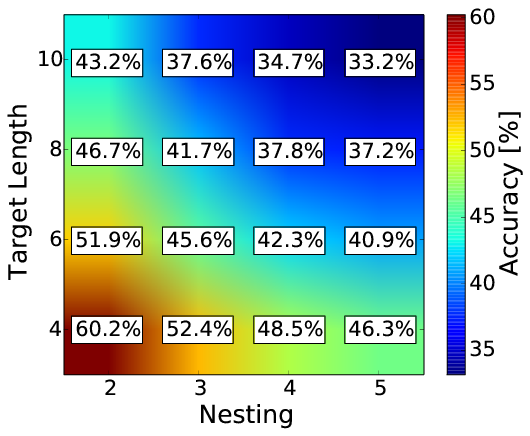}
    \end{minipage}
    \hfill
    \begin{minipage}[b]{0.32\textwidth}
        \centering
        \includegraphics[width=0.98\columnwidth]{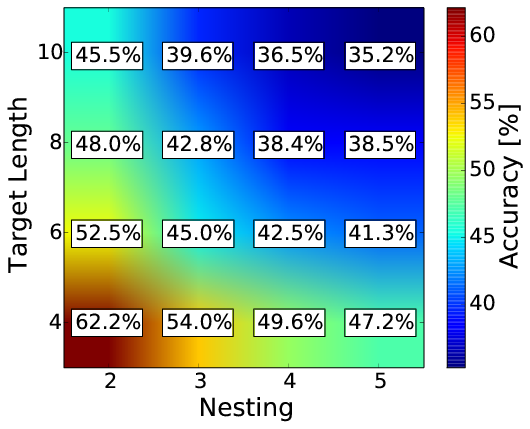}
    \end{minipage}
    \hfill
    \begin{minipage}[b]{0.32\textwidth}
        \centering
        \includegraphics[width=0.98\columnwidth]{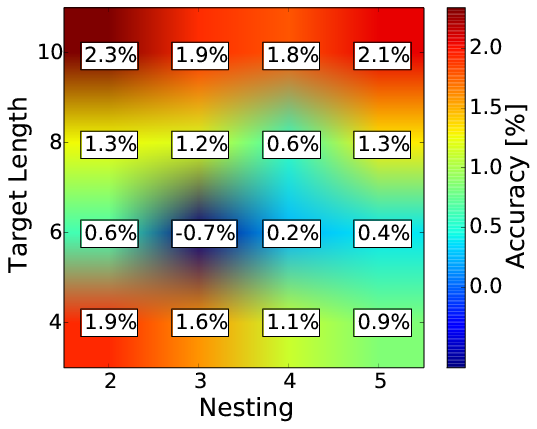}
    \end{minipage}

    \centering
    \begin{minipage}[b]{0.32\textwidth}
        \centering
        \includegraphics[width=0.98\columnwidth]{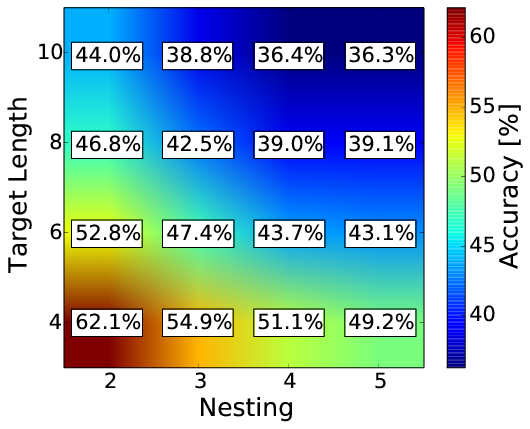}
    \end{minipage}
    \hfill
    \begin{minipage}[b]{0.32\textwidth}
        \centering
        \includegraphics[width=0.98\columnwidth]{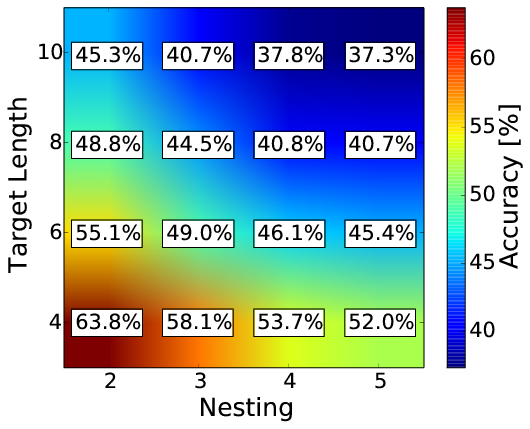}
    \end{minipage}
    \hfill
    \begin{minipage}[b]{0.32\textwidth}
        \centering
        \includegraphics[width=0.98\columnwidth]{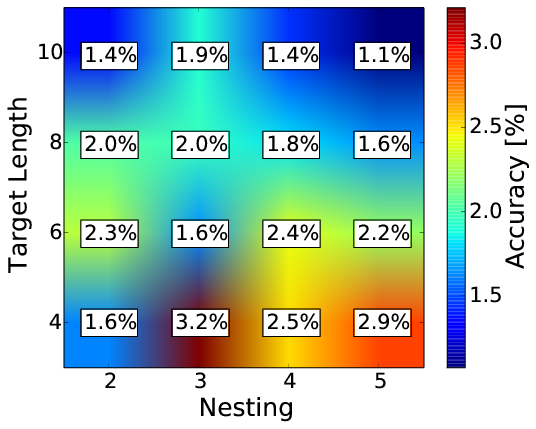}
    \end{minipage}

    \begin{minipage}{0.32\textwidth}
        \centering
        (a) Stacked RNN
    \end{minipage}
    \hfill
    \begin{minipage}{0.32\textwidth}
        \centering
        (b) Gated Feedback RNN
    \end{minipage}
    \hfill
    \begin{minipage}{0.32\textwidth}
        \centering
        (c) Gaps between (a) and (b)
    \end{minipage}

    \caption{
        Heatmaps of (a) Stacked RNN, (b) GF-RNN, and (c) difference obtained by
        substracting (a) from (b). The top row is the heatmaps of models using GRUs,
        and the bottom row represents the heatmaps of the models using LSTM units.
        Best viewed in colors.
    }
    \label{fig:lte_heatmaps}
\end{figure*}

\textbf{Large GF-RNN}

We trained a larger GF-RNN that has five recurrent layers, each of which has $700$
LSTM units. This makes it possible for us to compare the performance of the
proposed architecture against the previously reported results using other types
of RNNs. In Table~\ref{tab:hutter_results_large}, we present the test set BPC
by a multiplicative RNN~\citep{sutskever2011generating}, a stacked
LSTM~\citep{graves2013generating} and the GF-RNN with LSTM units.
The performance of the proposed GF-RNN is comparable to, or better than,
the previously reported best results. Note that
\citet{sutskever2011generating} used the vocabulary of 86 characters
(removed XML tags and the Wikipedia markups), and their result is not directly comparable with ours.
In this experiment, we used Adam %~\citep{kingma2014adam}
instead of RMSProp
to optimize the RNN. We used learning rate of $0.001$ and $\beta_1$ and $\beta_2$
were set to $0.9$ and $0.99$, respectively.

\subsection{Python Program Evaluation}

%change_5 Details on python program evaluation test procedure
%In Python program evaluation, we closely follow \citep{zaremba2014learning} and
%compute the test accuracy as the next step character prediction given a sequence
%of correct preceding characters.
%change_5 end

Fig.~\ref{fig:lte_heatmaps} presents the test results of each model represented
in heatmaps. The accuracy tends to decrease by the growth of the length of
target sequences or the number of nesting levels, where the difficulty or
complexity of the Python program increases.  We observed that in most of the
test sets, GF-RNNs are outperforming stacked RNNs, regardless of the type of
units.  Fig.~\ref{fig:lte_heatmaps}~(c) represents the gaps between the test
accuracies of stacked RNNs and GF-RNNs which are computed by subtracting (a)
from (b).  In Fig.~\ref{fig:lte_heatmaps}~(c), the red and yellow colors,
indicating large gains, are concentrated on top or right regions (either the
number of nesting levels or the length of target sequences increases). From this
we can more easily see that the GF-RNN outperforms the stacked RNN, especially
as the number of nesting levels grows or the length of target sequences
increases.

\section{Conclusion}
\label{sec:conclusion}

We proposed a novel architecture for deep stacked RNNs which uses gated-feedback
connections between different layers.  Our experiments focused on challenging
sequence modeling tasks of character-level language modeling and Python program
evaluation. The results were consistent over different datasets, and clearly
demonstrated that gated-feedback architecture is helpful when the models are
trained on complicated sequences that involve long-term dependencies.  We also
showed that gated-feedback architecture was faster in wall-clock time over the
training and achieved better performance compared to standard stacked RNN with a
same amount of capacity. Large GF-LSTM was able to outperform the previously
reported best results on character-level language modeling. This suggests that
GF-RNNs are also scalable. GF-RNNs were able to outperform standard stacked
RNNs and the best previous records on Python program evaluation task with
varying difficulties.

We noticed a deterioration in performance when the proposed
gated-feedback architecture was used together with a $\tanh$ activation
function, unlike when it was used with more sophisticated gated activation
functions. More thorough investigation into the interaction between the
gated-feedback connections and the role of recurrent activation function is
required in the future.

%GF-RNN is a simple extension over the conventional stacked RNNs, but it was able
%to show significant improvements on our benchmark and control experiments.

%\newpage
\section*{Acknowledgments}
\label{sec:ack}

The authors would like to thank the developers of
Theano~\citep{Bastien-Theano-2012} and
Pylearn2~\citep{pylearn2_arxiv_2013}. Also, the authors thank Yann N.
Dauphin and Laurent Dinh for insightful comments and discussion.  We acknowledge the support
of the following agencies for research funding and computing support: NSERC,
Samsung, Calcul Qu\'{e}bec, Compute Canada, the Canada Research Chairs and
CIFAR.

%\newpage
\bibliography{./bib/strings,./bib/strings-shorter,./bib/ml,./bib/aigaion,myref}

\begin{thebibliography}{25}
\providecommand{\natexlab}[1]{#1}
\providecommand{\url}[1]{\texttt{#1}}
\expandafter\ifx\csname urlstyle\endcsname\relax
  \providecommand{\doi}[1]{doi: #1}\else
  \providecommand{\doi}{doi: \begingroup \urlstyle{rm}\Url}\fi

\bibitem[Bahdanau et~al.(2014)Bahdanau, Cho, and
  Bengio]{Bahdanau-et-al-arxiv2014}
Bahdanau, Dzmitry, Cho, Kyunghyun, and Bengio, Yoshua.
\newblock Neural machine translation by jointly learning to align and
  translate.
\newblock Technical report, arXiv preprint arXiv:1409.0473, 2014.

\bibitem[Bastien et~al.(2012)Bastien, Lamblin, Pascanu, Bergstra, Goodfellow,
  Bergeron, Bouchard, and Bengio]{Bastien-Theano-2012}
Bastien, Fr{\'{e}}d{\'{e}}ric, Lamblin, Pascal, Pascanu, Razvan, Bergstra,
  James, Goodfellow, Ian~J., Bergeron, Arnaud, Bouchard, Nicolas, and Bengio,
  Yoshua.
\newblock Theano: new features and speed improvements.
\newblock Deep Learning and Unsupervised Feature Learning NIPS 2012 Workshop,
  2012.

\bibitem[Bengio et~al.(1994)Bengio, Simard, and Frasconi]{Bengio1994ITNN}
Bengio, Yoshua, Simard, Patrice, and Frasconi, Paolo.
\newblock Learning long-term dependencies with gradient descent is difficult.
\newblock \emph{IEEE Transactions on Neural Networks}, 5\penalty0 (2):\penalty0
  157--166, 1994.

\bibitem[Bengio et~al.(2001)Bengio, Ducharme, and Vincent]{BenDucVin01-short}
Bengio, Yoshua, Ducharme, R\'ejean, and Vincent, Pascal.
\newblock A neural probabilistic language model.
\newblock In \emph{Adv. Neural Inf. Proc. Sys. 13}, pp.\  932--938, 2001.

\bibitem[Cho et~al.(2014)Cho, Van~Merri\"enboer, Gulcehre, Bougares, Schwenk,
  and Bengio]{cho2014learning}
Cho, Kyunghyun, Van~Merri\"enboer, Bart, Gulcehre, Caglar, Bougares, Fethi,
  Schwenk, Holger, and Bengio, Yoshua.
\newblock Learning phrase representations using rnn encoder-decoder for
  statistical machine translation.
\newblock \emph{arXiv preprint arXiv:1406.1078}, 2014.

\bibitem[El~Hihi \& Bengio(1995)El~Hihi and Bengio]{el1995hierarchical}
El~Hihi, Salah and Bengio, Yoshua.
\newblock Hierarchical recurrent neural networks for long-term dependencies.
\newblock In \emph{Advances in Neural Information Processing Systems}, pp.\
  493--499. Citeseer, 1995.

\bibitem[Gers et~al.(2000)Gers, Schmidhuber, and Cummins]{Gers2000}
Gers, Felix~A., Schmidhuber, J{\"{u}}rgen, and Cummins, Fred~A.
\newblock Learning to forget: Continual prediction with {LSTM}.
\newblock \emph{Neural Computation}, 12\penalty0 (10):\penalty0 2451--2471,
  2000.

\bibitem[Goodfellow et~al.(2013)Goodfellow, Warde-Farley, Lamblin, Dumoulin,
  Mirza, Pascanu, Bergstra, Bastien, and Bengio]{pylearn2_arxiv_2013}
Goodfellow, Ian~J., Warde-Farley, David, Lamblin, Pascal, Dumoulin, Vincent,
  Mirza, Mehdi, Pascanu, Razvan, Bergstra, James, Bastien,
  Fr{\'{e}}d{\'{e}}ric, and Bengio, Yoshua.
\newblock Pylearn2: a machine learning research library.
\newblock \emph{arXiv preprint arXiv:1308.4214}, 2013.

\bibitem[Graves(2013)]{graves2013generating}
Graves, Alex.
\newblock Generating sequences with recurrent neural networks.
\newblock \emph{arXiv preprint arXiv:1308.0850}, 2013.

\bibitem[Hermans \& Schrauwen(2013)Hermans and Schrauwen]{hermans2013training}
Hermans, Michiel and Schrauwen, Benjamin.
\newblock Training and analysing deep recurrent neural networks.
\newblock In \emph{Advances in Neural Information Processing Systems}, pp.\
  190--198, 2013.

\bibitem[Hinton(2012)]{Hinton-Coursera2012}
Hinton, Geoffrey.
\newblock Neural networks for machine learning.
\newblock Coursera, video lectures, 2012.

\bibitem[Hochreiter(1991)]{Hochreiter91}
Hochreiter, Sepp.
\newblock { Untersuchungen zu dynamischen neuronalen Netzen. Diploma thesis,
  Institut f\"{u}r Informatik, Lehrstuhl Prof. Brauer, Technische
  Universit\"{a}t M\"{u}nchen}, 1991.
\newblock URL \url{http://www7.informatik.tu-muenchen.de/~Ehochreit}.

\bibitem[Hochreiter(1998)]{hochreiter1998vanishing}
Hochreiter, Sepp.
\newblock The vanishing gradient problem during learning recurrent neural nets
  and problem solutions.
\newblock \emph{International Journal of Uncertainty, Fuzziness and
  Knowledge-Based Systems}, 6\penalty0 (02):\penalty0 107--116, 1998.

\bibitem[Hochreiter \& Schmidhuber(1997)Hochreiter and
  Schmidhuber]{Hochreiter+Schmidhuber-1997}
Hochreiter, Sepp and Schmidhuber, J{\"u}rgen.
\newblock Long short-term memory.
\newblock \emph{Neural Computation}, 9\penalty0 (8):\penalty0 1735--1780, 1997.

\bibitem[Hutter(2012)]{Hutter2012}
Hutter, Marcus.
\newblock The human knowledge compression contest.
\newblock 2012.
\newblock URL \url{http://prize.hutter1.net/}.

\bibitem[Kingma \& Ba(2014)Kingma and Ba]{kingma2014adam}
Kingma, Diederik and Ba, Jimmy.
\newblock Adam: A method for stochastic optimization.
\newblock \emph{arXiv preprint arXiv:1412.6980}, 2014.

\bibitem[Koutn{\'\i}k et~al.(2014)Koutn{\'\i}k, Greff, Gomez, and
  Schmidhuber]{koutnik2014clockwork}
Koutn{\'\i}k, Jan, Greff, Klaus, Gomez, Faustino, and Schmidhuber, J{\"u}rgen.
\newblock A clockwork rnn.
\newblock In \emph{Proceedings of the 31st International Conference on Machine
  Learning (ICML'14)}, 2014.

\bibitem[Mikolov(2012)]{Mikolov-thesis-2012}
Mikolov, Tomas.
\newblock \emph{Statistical Language Models based on Neural Networks}.
\newblock PhD thesis, Brno University of Technology, 2012.

\bibitem[Mikolov et~al.(2012)Mikolov, Sutskever, Deoras, Le, Kombrink, and
  Cernocky]{mikolov2012subword}
Mikolov, Tomas, Sutskever, Ilya, Deoras, Anoop, Le, Hai-Son, Kombrink, Stefan,
  and Cernocky, J.
\newblock Subword language modeling with neural networks.
\newblock \emph{Preprint}, 2012.

\bibitem[Schmidhuber(1992)]{Schmidhuber92}
Schmidhuber, J{\"u}rgen.
\newblock Learning complex, extended sequences using the principle of history
  compression.
\newblock \emph{Neural Computation}, 4\penalty0 (2):\penalty0 234--242, 1992.

\bibitem[Stollenga et~al.(2014)Stollenga, Masci, Gomez, and
  Schmidhuber]{stollenga2014deep}
Stollenga, Marijn~F, Masci, Jonathan, Gomez, Faustino, and Schmidhuber,
  J{\"u}rgen.
\newblock Deep networks with internal selective attention through feedback
  connections.
\newblock In \emph{Advances in Neural Information Processing Systems}, pp.\
  3545--3553, 2014.

\bibitem[Sutskever et~al.(2011)Sutskever, Martens, and
  Hinton]{sutskever2011generating}
Sutskever, Ilya, Martens, James, and Hinton, Geoffrey~E.
\newblock Generating text with recurrent neural networks.
\newblock In \emph{Proceedings of the 28th International Conference on Machine
  Learning (ICML'11)}, pp.\  1017--1024, 2011.

\bibitem[Sutskever et~al.(2014)Sutskever, Vinyals, and
  Le]{sutskever2014sequence}
Sutskever, Ilya, Vinyals, Oriol, and Le, Quoc~VV.
\newblock Sequence to sequence learning with neural networks.
\newblock In \emph{Advances in Neural Information Processing Systems}, pp.\
  3104--3112, 2014.

\bibitem[Zaremba \& Sutskever(2014)Zaremba and Sutskever]{zaremba2014learning}
Zaremba, Wojciech and Sutskever, Ilya.
\newblock Learning to execute.
\newblock \emph{arXiv preprint arXiv:1410.4615}, 2014.

\bibitem[Zaremba et~al.(2014)Zaremba, Sutskever, and Vinyals]{zaremba2014}
Zaremba, Wojciech, Sutskever, Ilya, and Vinyals, Oriol.
\newblock Recurrent neural network regularization.
\newblock \emph{arXiv preprint arXiv:1409.2329}, 2014.

\end{thebibliography}
%\bibliography{strings,strings-shorter,ml,aigaion,myref}
%\bibliography{strings,strings-shorter,ml,aigaion,junyoung}
%\bibliographystyle{../../bst/icml2015}
\bibliographystyle{./template/icml2015}

%\newpage
%\appendix
%\twocolumn[
%\icmltitle{Supplementary Material: Gated Feedback Recurrent Neural Networks}
%\vskip 0.3in
%]
%\input{supp.tex}

\end{document}